\documentclass{llncs}

\usepackage{cite}
\usepackage[pdftex]{graphicx}
\usepackage{amsmath}
\usepackage[caption=false]{subfig}
\usepackage{hyperref}
\usepackage{textcomp}

\usepackage{pgfplots}
\pgfplotsset{compat=1.12}
\newlength\figureheight
\newlength\figurewidth

\begin{document}

© 2022  IAS Society. Personal use of this material is permitted.  Permission from IAS Society must be obtained for all other uses, in any current or future media, including reprinting/republishing this material for advertising or promotional purposes, creating new collective works, for resale or redistribution to servers or lists, or reuse of any copyrighted component of this work in other works.

\newpage

\title{Gestural and Touchscreen Interaction for Human-Robot Collaboration: a Comparative Study}

\author{Antonino Bongiovanni \thanks{The authors contributed equally.}, Alessio De Luca $^*$, Luna Gava $^*$, Lucrezia Grassi $^*$, Marta Lagomarsino$^*$, Marco Lapolla $^*$, Antonio Marino $^*$, Patrick Roncagliolo $^*$, Simone Macci\`o, Alessandro Carf\`i, and Fulvio Mastrogiovanni
}
\institute{Department of Informatics, Bioengineering, Robotics, and Systems Engineering, University of Genoa, Via Opera Pia 13, 16145 Genoa, Italy \\
\email{alessandro.carfi@dibris.unige.it}}

\maketitle

\begin{abstract}
Close human-robot interaction (HRI), especially in industrial scenarios, has been vastly investigated for the advantages of combining human and robot skills. For an effective HRI, the validity of currently available human-machine communication media or tools should be questioned, and new communication modalities should be explored. This article proposes a modular architecture allowing human operators to interact with robots through different modalities. In particular, we implemented the architecture to handle gestural and touchscreen input, respectively, using a smartwatch and a tablet. Finally, we performed a comparative user experience study between these two modalities.
\end{abstract}

\section{Introduction}
\label{sec:introduction}

Research on human-robot collaboration (HRC) leapt forward in industrial scenarios with the introduction of collaborative robots (cobots) \cite{colgate1996cobots} such as the Kuka LBR iiwa, the Universal Robot UR5, or ABB YuMi (to mention just a few). Cobots are safe by design and make it possible for robots and humans to combine their skills, improving overall productivity, efficiency and flexibility while, possibly, reducing human stress and workload \cite{Tsarouchi2016}. For a functional human-robot collaboration, safety is not the only requirement, and cobots should provide more intuitive interfaces moving away from classical ones such as teach pendants.

More prominent human-robot interfaces rely on touchscreen technology \cite{Mateo2014, birkenkampf2014knowledge} that integrates into the same device the human to robot communication (touch) and the robot to human one (screen). Touch screen interfaces are classical graphical user interfaces (GUIs) where the presence of the physical screen acts as a gateway between the human and the robot. New technologies such as augmented and virtual reality provide alternatives for robots to communicate with humans. These interfaces need novel approaches for human interaction, and researchers took inspiration from human-human communication exploring speech \cite{poirier2019voice}, gestures \cite{neto2019gesture, Cicirelli2015, carfi2021gesture} and their combination \cite{papanastasiou2019towards, wu2020development, gromov2016wearable}.

Gestural- and speech-based interfaces are appealing since humans already use these modalities to interact with other humans. However, their usage in an intrinsically different scenario may lead to sub-optimal interaction results. For these reasons, researchers performed experiments to compare new interaction modalities with state-of-the-art ones, such as those based on touchscreens. Often, the number of participants of these studies is not significant, and they rely on Wizard of Oz (WoZ) experiments (i.e., experiments in which the subject interacts with an autonomous system, but in reality, an experimenter is operating it) \cite{pohlt2018effects}.

\begin{figure}[t]
    \centering
    \subfloat[]
    [Touchscreen interface.]
    {\includegraphics[width=0.8\linewidth]{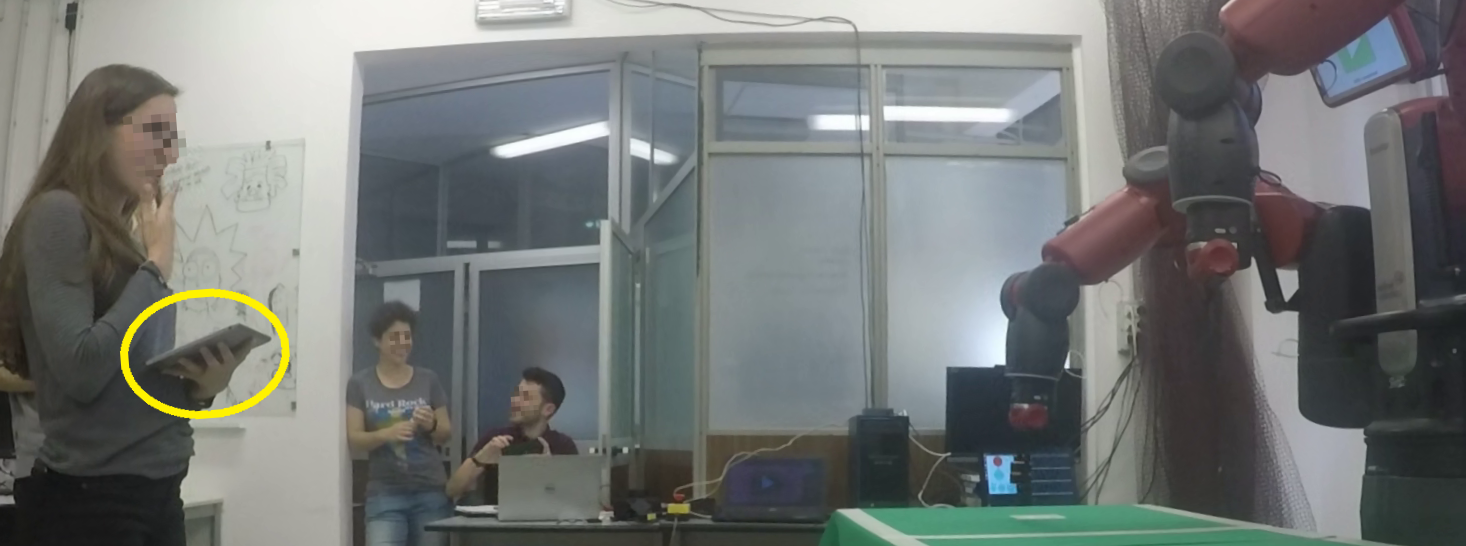}\label{fig:tablet}}\\
    \subfloat[]
    [Gesture-based interface.]
    {\includegraphics[width=0.8\linewidth]{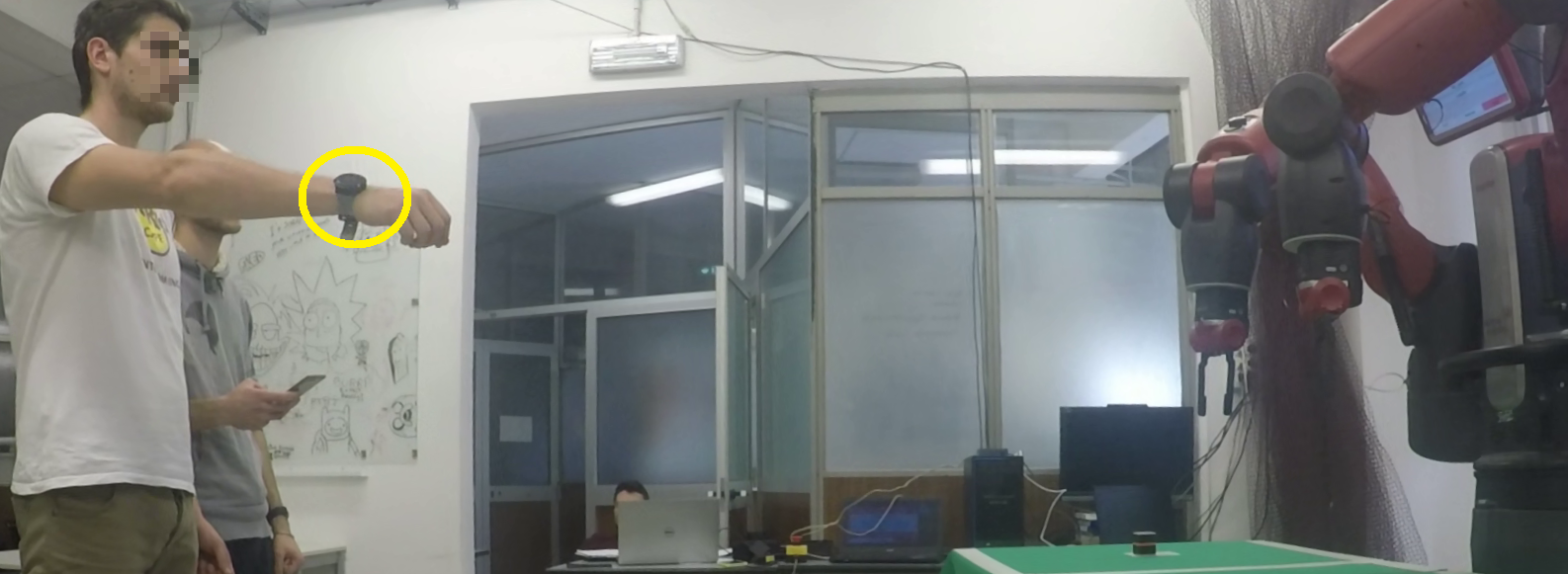}\label{fig:smartwatch}}
    \caption{Two users using, respectively, a tablet and a smartwatch,  highlighted by yellow circles, to interact with a robot.}
    \label{fig:examples}
\end{figure}

Since previous studies highlighted them as one of the most promising approaches, we aim to contribute to this research field by evaluating the user experience and performance results of a gesture-based interface for HRC. We do this by comparing gestural (Figure \ref{fig:smartwatch}) with touchscreen (Figure \ref{fig:tablet}) interaction in a real-world HRC scenario. Touchscreen interaction has been picked as an optimal term of comparison because of its modern and reliable technology.
For the experiment, although it could be suboptimal in the HRC scenario, we adopted a simple GUI to have optimal conditions for the touchscreen interface since a GUI is part of its usual setup. Part of the contribution of this work is the design and implementation of a software architecture capable of handling both communication modalities.


The paper is structured as follows. In Section \ref{sec:background} a brief review of HRC interfaces literature is presented. Section \ref{sec:principles} describes the design principles adopted to design our software architecture. Section \ref{sec:implementation} describes, in detail, the architecture implementation for the gestural and touchscreen interaction and the logic managing the HRC. The setup and the experiment description are presented in Section \ref{sec:experiment}, while in Section \ref{sec:result} the experimental results are presented and discussed. Conclusions follow.

\section{Background}
\label{sec:background}

While humans and robots collaborate, they share information at different levels through explicit or implicit communication. Ideally, an interface for human-robot collaboration should mediate the two-way communication handling both explicit and implicit communication, but in this paper, we focus only on the explicit one. In this context, the interface allows the human to send direct commands to the robot and the robot answers providing appropriate feedback. Classical interfaces use screens to give the user visual feedback but, novel technologies, such as augmented reality (AR) and see-through displays \cite{Taralle2015, Ziegler2015, Aromaa2016}, allow integrating the interface directly in the working environment. 

At the same time, researchers explored different modalities to acquire human commands. Many studies focused on touchscreen interfaces for human-robot interaction, given their widespread adoption in the consumer market. Mateo et al. 2014 \cite{Mateo2014} proposed a tablet-based user interface for industrial robot programming, and a similar interface has been used to send high-level commands to robotic platforms \cite{birkenkampf2014knowledge}. However, touchscreen interfaces are suboptimal in scenarios where humans need free hands. Furthermore, the need for a screen to communicate limits the adoption of technologies such as AR that can only be used as a complement \cite{papanastasiou2019towards}, and not fully substitute it. Therefore, developing new interfaces requires alternative communication modalities such as speech and gestures. Despite the extended research in speech analysis and recognition \cite{novoa2018}, the high noise level in industrial environments can jeopardize the correct function of speech-based interfaces. While gesture-based interfaces usually rely on upper-limb motions, possibly altering the human workflow. However, with gesture-based interfaces, users can send commands to the robot while handling tools; and without using a tablet or reaching a screen. Furthermore, interfaces can rely on gestures for scenarios where other communication modalities fail, e.g., underwater HRC \cite{Islam2019}.

Because of their advantages, researchers have explored the usage of gesture-based interfaces for industrial applications \cite{neto2019gesture}, search and rescue \cite{gromov2016wearable}, and direct control of mobile robots \cite{Cicirelli2015}. Neto et al. 2019 \cite{neto2019gesture} proposed a gesture-based interface using five inertial measurement units (IMUs), worn by the user, to send commands to an industrial manipulator. At the same time, other studies proposed interfaces using non-wearable devices, such as RGB \cite{Islam2019 } and RGB-D \cite{Cicirelli2015} cameras, or alternative wearable devices such as Electromyography (EMG) sensors \cite{gromov2016wearable, Ahsan2011}. Although wearable devices can impede human motions and rely on batteries, their usage reduces the necessity for a structured environment, raises fewer privacy concerns, and does not suffer occlusions. IMUs are one of the most used sensing solutions for wearables because of their compact data stream, reduced price and small size. Independently by the sensor type, a gesture-based interface should process the collected data to identify motions with explicit communication intent and ignore normal user operations. This problem is known as gesture recognition, and for IMU data, it has been approached using Naive Bayes Classifiers, Logistic Regression and Decision Trees \cite{Xu2015}, Convolutional Neural Networks (CNNs) \cite{Kwon2018}, Dynamic Time Warping (DTW) \cite{Yu-LiangHsu2015}, Support Vector Machines (SVMs) \cite{Wen2016}, and Long Short-Term Memory Neural Networks (LSTM) \cite{Carfi2018}.


\begin{figure}[t]
    \centering
    \includegraphics[width=\linewidth]{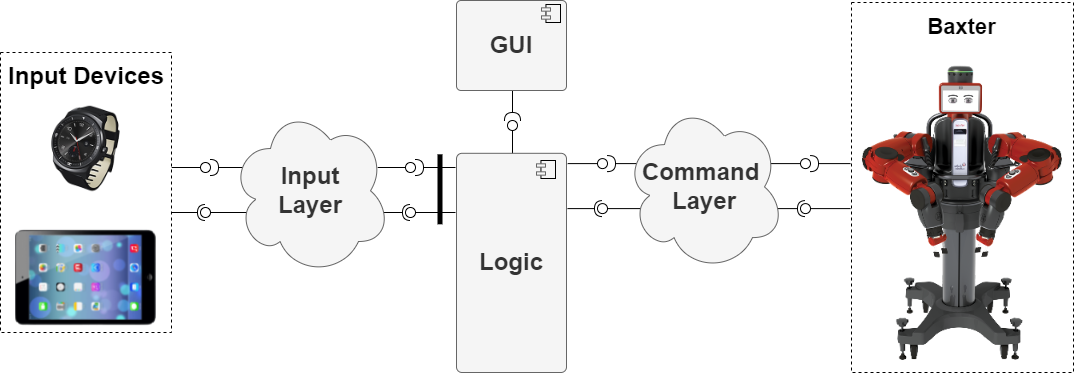}
    \caption{A conceptual flow diagram of the proposed architecture.}
    \label{fig:GEN_arch}
\end{figure}

The development of a new interface is not only a technological problem but, the human factor should be taken into consideration as well \cite{wachs2011vision}. Therefore, an experimental validation should evaluate both the system performance and the user experience. For this reason, some studies proposed comparative analyses between different interaction modalities to determine which is the most suited given a target application. Taralle et al. 2015 \cite{Taralle2015} presented a comparative study between a gesture-based and touchscreen interface to control a small unmanned aerial vehicle (sUAV) while looking for a target in a video stream. This study relied on WoZ experiments for the gesture controls, and the results, over 20 volunteers, suggested a preference for the gestural interaction. At the same time, lower cognitive load and user preference for gestural and touch inputs have been found using WoZ experiments to compare touch, gesture, speech, and 3D pen in an industrial scenario \cite{profanter2015analysis}. Although these studies provide promising results, WoZ experiments overlook the effect of the system performances (e.g., accuracy, responsiveness) on the user experience. When compared in a real-world environment, reliable input modalities, such as touchscreens, outperform novel ones, such as gestures \cite{pohlt2018effects}. These results should not discourage the study of new interaction modalities but rather push to reach better system performances to i) obtain interfaces for a more flexible human-robot collaboration and ii) overcome the need for a screen in classical interfaces.

\begin{figure}[t]
    \centering
    \vspace{0.2cm}
    \includegraphics[width=0.65\linewidth]{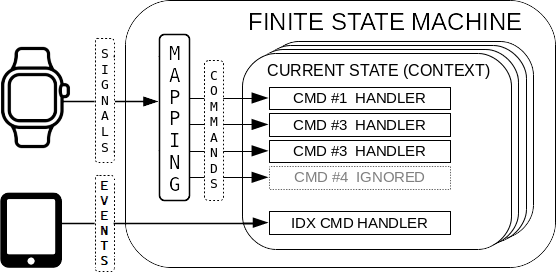}
    \caption{Many heterogeneous discrete \textit{signals} are mapped to one of the four FSM command handlers. Each FSM state can install the handlers, or ignore them (like command number 4 in this example). A secondary input scheme, i.e., \textit{events}, uses valued signals (integers) to directly select one of the options offered by the FSM state for user selection.}
    \label{fig:HRI_schema}
\end{figure}

\section{Design principles}
\label{sec:principles}

As we have seen, an interface mediates the interaction between the human and the robot, allowing the human to send commands and provide appropriate feedback. In Figure \ref{fig:GEN_arch}, a general architecture summarizing this concept is presented. The architecture consists of: a GUI displayed on a screen, an input layer, the main logic, and a command layer that communicates directly with the robot. Since our study focuses on the human to robot communication, we kept the robot to human one simple using only a screen. Since they are vastly adopted, we design a menu-based GUI listing all possible functionalities to minimize the novelty effect. Furthermore, menus allow an easy adaptation to different experimental scenarios. 

The input layer collects the sensory information and processes them to recognize discrete human commands (i.e., arm gestures, keywords, or touchscreen pressures). Each communication modality has its dictionary $D$ containing commands descriptions and associated identifiers. Once the sensory information matches the command description, the input layer returns as output the associated identifier. In some communication modalities, the input layer also generates additional information. For example, when the user presses a touchscreen, the input layer returns the identifier (i.e., screen pressure) and the related 2D position.

The logic layer is represented using a finite state machine (FSM) whose states describe different interaction stages. The set of states $S$ ($S=\{s_1, \dots, s_i, \dots s_{|S|}\}$) composes the FSM, and the transitions from $s_i$ ($s_i \in S$) to other states are described by the transitions set $T_i$. In our scenario, we recognize two states categories, namely \textit{menu} and \textit{action}. Menu states describe the GUI (Figure \ref{fig:menu_states} shows the GUI representation of some menu states), and when they are active, the interaction is limited to the menu navigation. Instead, action states implement the system functionalities, e.g., human teaching of a motion or robot execution of a task.

Transitions between FSM states are triggered either by the system (e.g., when the robot ends a task execution) or by human commands. As we have seen, different communication modalities can carry different information. Therefore, to make it possible to integrate various communication modalities, we designed two different schemas that a human command can follow to interact with the FSM: \textit{signals} and \textit{events} (see Figure \ref{fig:HRI_schema}). With signals, for each state $s_i$, the system defines a one-to-one mapping between the dictionary commands and the transitions defined in $T_i$. Command handlers, defined for each FSM state (see Figure \ref{fig:HRI_schema}), manage this mapping. With events, human commands are mapped directly to FSM states. Therefore, when the human performs the command associated with $s_j$, independently by the current FSM state, $s_j$ is activated. This schema allows long jumps in the interface without a sequential transition through all the intermediate states. Usually, the number of states $|S|$ is higher than the number of transitions for a single state $|T_i|$, therefore using events needs bigger dictionaries or communication modalities that, carrying extra information, allow to associate more states to a single command. Signals are more appropriate for communication modalities where a big dictionary implies high user effort to recall commands (e.g., keyboard strokes or gestures). Instead, events can leverage communication modalities handling big dictionaries (e.g., vocal interaction) or providing extra information (e.g., touchscreen interaction). 


\section{Implementation}
\label{sec:implementation}
Following the design principles presented in the previous section, we have implemented our architecture (available open-source\footnote{Web: \url{https://github.com/TheEngineRoom-UniGe/gesture_based_interface}}) to interact with the dual-arm Baxter robot from Rethink Robotics \cite{Guizzo2012} using gestures, sensed by a wrist-worn IMU and a touchscreen. The architecture uses the robot operating system (ROS) framework \cite{Quigley2009} to manage inter-module communication. Besides benefits in code reuse, this choice allows us to exploit the Baxter-related ROS APIs, which exposes services to acquire sensory data, control robot actuators, and record robot motions using Kinesthetic Teaching (KT) \cite{kormushev2011, Carfi2019}.

\subsection{Graphical User Interface}
Our Graphical User Interface arranges the menu’s options vertically, and a red selector is used to highlight the selected option (see Figure \ref{fig:main_menu}). The FSM menu states describing the GUI are published on a ROS topic and converted, by a renderer, to a graphical representation. The GUI can be visualized on a screen either locally or remotely through a browser. Since the architecture decouples the GUI state representation (FSM) and its visual representation, it is possible to change the GUI appearance (e.g., textual user interfaces or GUI with different designs) or integrate other visualization devices (e.g., augmented reality headsets).

\begin{figure}[t]
    \centering
    \includegraphics[width=\linewidth]{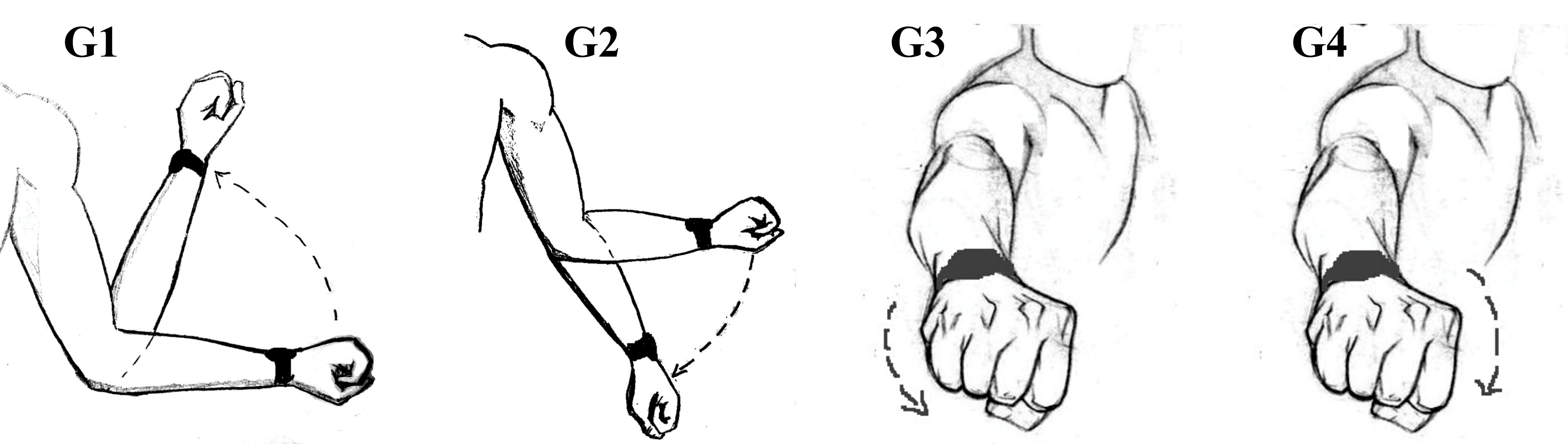}
    \caption{The four gestures supported by our architecture: (G1) wrist up, (G2) wrist down, (G3) spike clockwise, and (G4) spike counter-clockwise.}
    \label{fig:gestures}
\end{figure}


\subsection{Input}
The two interaction modalities considered in this study are touchscreen and gestures. The two modalities are intrinsically different, and for the reasons previously presented, we use signals to handle gestures and events for the touchscreen. While using gestures, users navigate the menus sequentially, performing multiple gestures to reach and select an option. With the touchscreen instead, the user should press simply the corresponding virtual button in the GUI. In our experiment, the touchscreen interface is used on an iPad Air through Safari, and we developed a ROS node receiving the command \textit{events} directly from the tablet.

For the gesture-based interface instead, we used SLOTH, a method taking advantage of LSTM neural networks to recognize gestures online. This method has been proposed for HRC scenarios and a ROS-compatible version, with a pre-trained neural network, is available on GitHub\footnote{Web: \url{https://github.com/ACarfi/SLOTH}}. We developed an Android Wear 2 application to collect IMU data from an LG G Watch R smartwatch and publish them online to a ROS topic. Therefore, since the data acquisition pipeline differs from the one used in the original work, the SLOTH performances should be assessed again \cite{Stisen2015}. Every time a gesture is recognized, the SLOTH module sends a \textit{signal} to the main logic activating an FSM transition. In our FSM implementation, the maximum number of possible transitions associated with a state is four ($\forall i \in \{1, \dots, |S|\}, |T_i| \leq 4$), therefore the architecture presents only four command handlers (see Figure \ref{fig:HRI_schema}). Since SLOTH recognizes between six gestures ($|D| = 6$), we ignore two of them. The command handlers map the others (see Figure \ref{fig:gestures}) as follows: command handler \#1 maps wrist up (G1), command handler \#2 maps wrist down(G2), command handler \#3 maps spike clockwise (G3), and command handler \#4 maps spike counter-clockwise (G4). Notice that the number of active command handlers in a state $s_i$ corresponds to the number of possible transition $T_i$, and gestures mapped by inactive command handlers would be effectless.

\subsection{Logic}
\begin{figure*}[t]
    \centering
    \subfloat[][Main]{\includegraphics[width=0.32\linewidth]{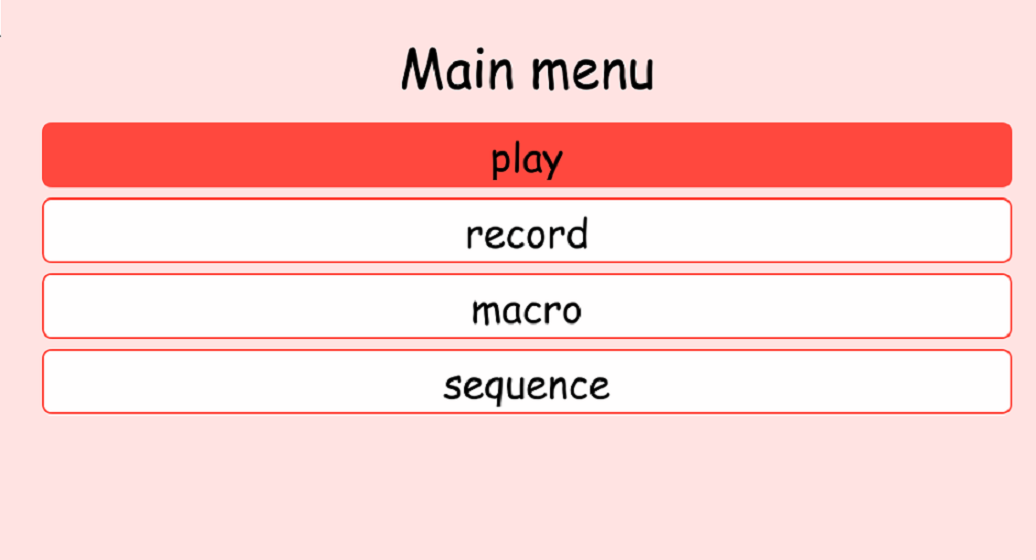}\label{fig:main_menu}}\hspace{0.01cm}
    \subfloat[][Record]{\includegraphics[width=.32\linewidth]{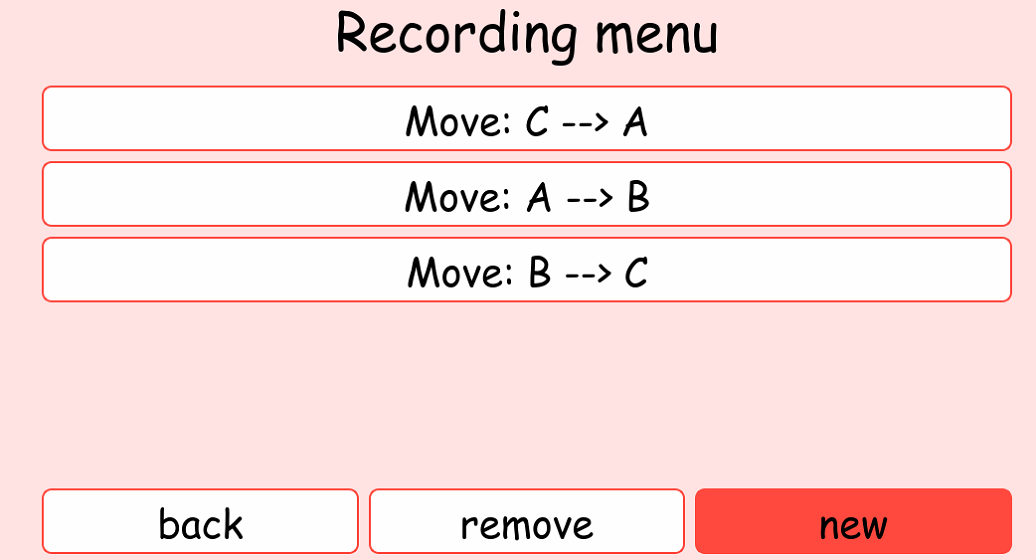}\label{fig:HRI_record}} \hspace{0.01 cm}
    \subfloat[][Playback]{\includegraphics[width=.32\linewidth]{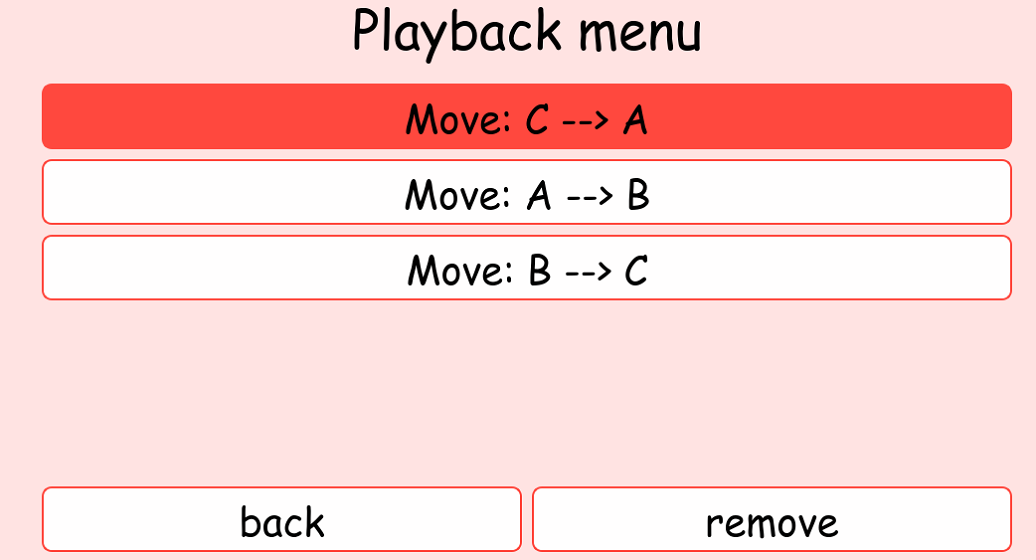}\label{fig:HRI_play}} 
    \\
    \subfloat[][Sequential playback]{\includegraphics[width=.32\linewidth]{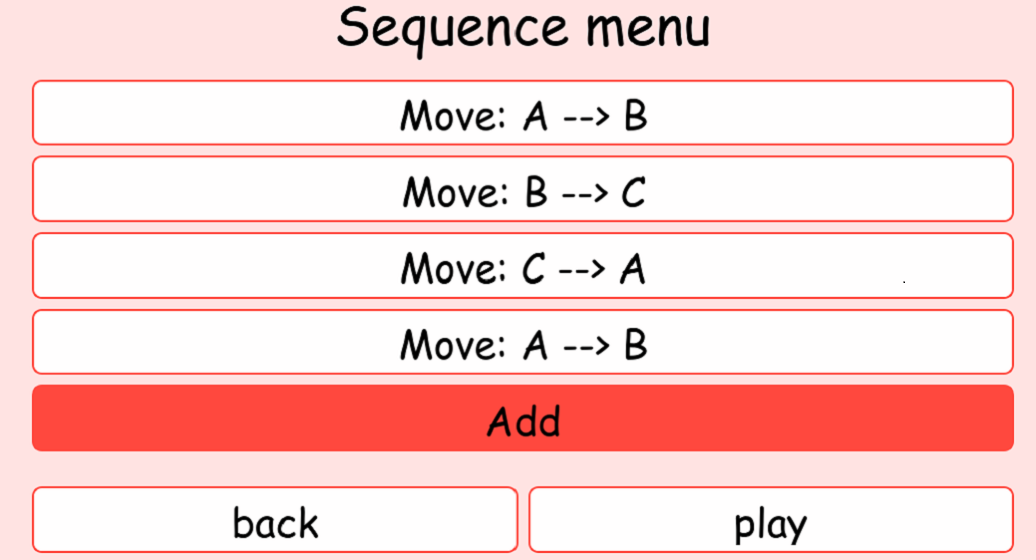}\label{fig:HRI_sequence}}\hspace{0.01 cm}
    \subfloat[][Macro mode]{\includegraphics[width=.32\linewidth]{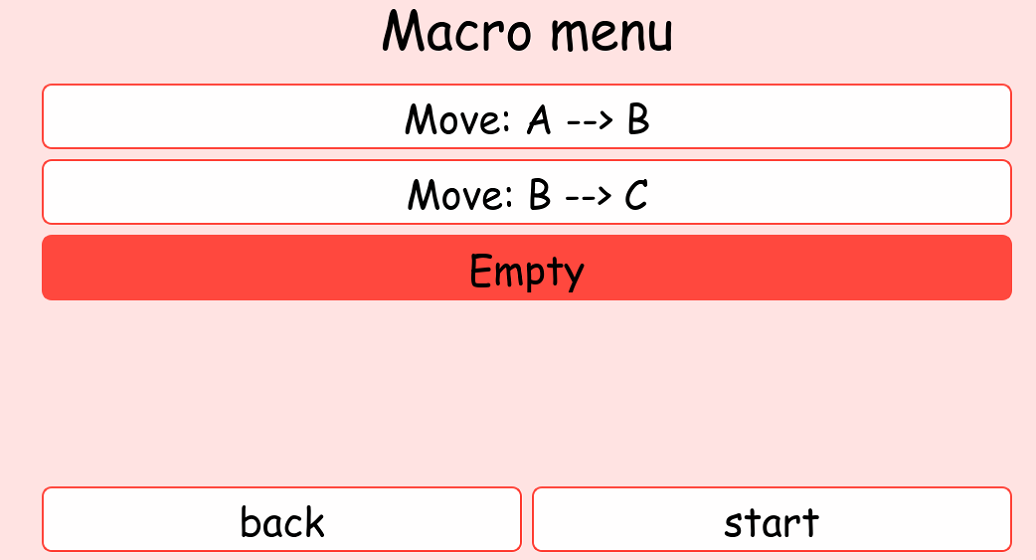}\label{fig:HRI_macro}}
    \caption{The menus for the four options provided by the interface.}
    \label{fig:menu_states}
\end{figure*}
As previously described, we implemented the architecture logic using an FSM that the user can control using signals or events. When using the gesture interface, each gesture execution generates a signal triggering the corresponding state transition. In this case, whichever change in the GUI (e.g., selector up, selector down or option selection) corresponds to a state transition in the FSM. Users can navigate menu states using three gestures: G1 to select options, G2 to move down the selector, and G3 to move up the selector (in menu states G4 is effectless). 

Our architecture allows the user to control robot functionalities to record and playback end-effector trajectories. In particular, the main menu offers four options: \textit{record}, \textit{playback}, \textit{sequential playback}, and \textit{macro mode} (see Figure \ref{fig:main_menu}). Whenever one of these options is selected, the corresponding menu is opened.

The record menu (Figure \ref{fig:HRI_record}) displays a list of recorded robot tasks (i.e., end-effector trajectories). The user can overwrite each task by selecting it or creating a new one using the corresponding option. By selecting a task from the list, the recording process, handled by the \textit{record} action state, starts. In this state, a human operator can program a new task using KT (i.e., teaching a trajectory by physically guiding the robot) and use G4 to terminate the recording going back to the record menu.

The playback menu (Figure \ref{fig:HRI_play}) lists all the saved tasks, which can be deleted by the human operator using the corresponding option. When a task is selected, the FSM transits to the \textit{playback action} state, and the robot reproduces the associated trajectory. Here two commands are active: G2 pausing or resuming the task playback and G4 terminating the playback and moving back to the playback menu.

The sequential playback menu (Figure \ref{fig:HRI_sequence}) allows the user to combine saved tasks in sequences to handle more complex behaviours. The user can remove or substitute a task from the sequence by selecting it. Furthermore, the \textit{add} option brings to a selection sub-menu where the operator can append a task to the sequence, and the \textit{run} option makes the FSM transit to the \textit{playback action} state that reproduces the whole sequence. 

Finally, the macro mode menu, see Figure \ref{fig:HRI_macro}, allows the user to associate one task to each of the three gestures G1, G2 and G3 (the correspondence task-gesture is ordered from top to bottom). The user can customize the mapping by selecting a specific slot and choosing a task from the list of available ones. When \textit{run} is selected, the FSM transits to the \textit{macro action} state, and the robot starts waiting for gestures. Every time the user performs one of the three mapped gestures, the robot executes the associated task. In this state, performing G4 terminates the interaction and moves back to the macro mode menu.

The logic does not vary when using the touchscreen interface. Instead of performing gestures, the user presses the touchscreen in a 2D position corresponding to a menu option, the system sends an event to the FSM, directly activating the corresponding state. 

\section{Experiment}
\label{sec:experiment}
Now that we have presented our architecture for multimodal human-robot interaction, we put forward our hypothesis inspired by previous researches. [Hypothesis $H_1$] the gesture-based interface will get positive user evaluation, and [$H_{1.1}$] the novelty effect will have a significant contribution. However, [$H_2$] the touchscreen interface will grant an overall better user experience [$H_{2.1}$] since it relies on a more stable technology and volunteers are accustomed to this kind of interface.

\subsection{Experimental Setup}

\begin{figure}[t]
    \centering
    \includegraphics[width=0.75\linewidth]{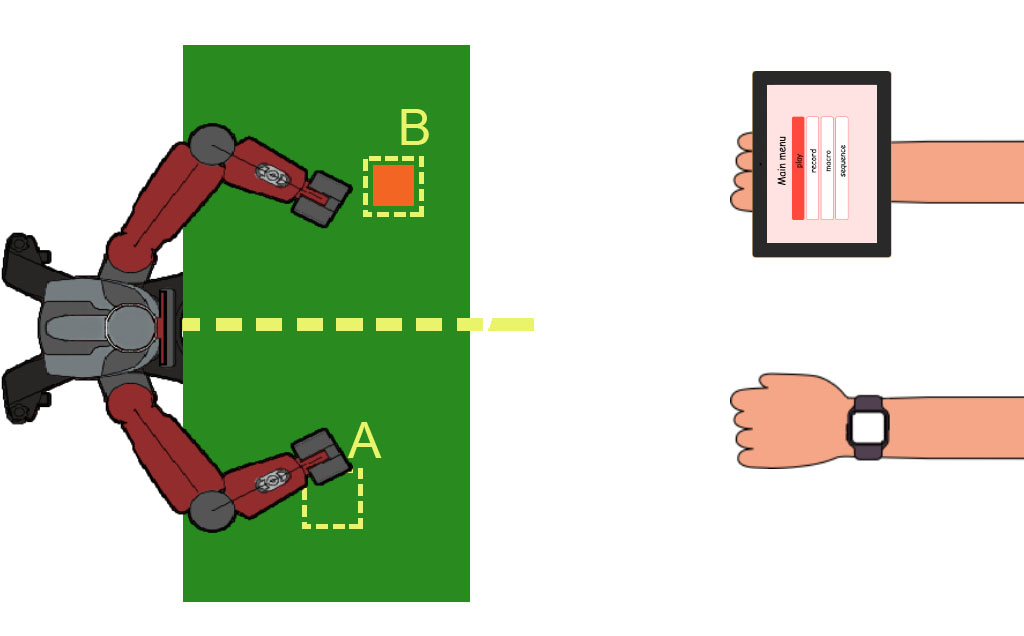}
    \caption{Experimental scenario: a human holds the tablet or wears the smartwatch, while Baxter is behind a table where the pick-and-places of a wooden cube are performed.}
    \label{fig:exp_setup}
\end{figure}

The experimental setup, whose conceptual sketch is represented in Figure \ref{fig:exp_setup}, includes:
(i) a dual-arm Baxter manipulator, appropriately configured, 
(ii) a wooden, $3.7$ $cm$ edge cube, to be manipulated by the robot, 
(iii) the tablet or the smartwatch that human operators use,
(iv) a table in front of the robot, on which two positions, namely $A$ and $B$, are defined. 
Position $A$ is where the cube is located at the beginning of the experiment.
The distance between $A$ and $B$ is 60 $cm$. 
While performing the experiments, the human volunteer stands in front of the robot at a distance of approximately $1.5$ $m$.

The architecture contains three pre-recorded tasks, namely $\textsc{Move: A} \rightarrow \textsc{B}$, \textsc{action 1}, reproducing a robot arm-waving to greet the user,  and \textsc{action 2}, performing a handover of the cube between the two Baxter's arms. These have been included to simulate the difficulty that using an already setup interface may imply.

\subsection{Description of the Experiment}

We performed a within-subject experiment divided into two phases, i.e., a trial using the tablet (touchscreen-based) and one wearing the smartwatch (gesture-based). The $25$ participants, in a range between $18$ and $40$ years old, have been divided into two groups. The two groups completed the experiment once with each device: the first group started using the smartwatch, and the second group with the tablet.

The experimental protocol consists of a series of transportation tasks of the wooden block from position $A$ to position $B$ and \textit{vice versa}. The success in bringing back the cube from $B$ to $A$ and the overall quality of the robot motions are not evaluation criteria considered in this experiment. Instead, as previously anticipated, we keep track of the time each operator needs to complete the experiment. Before the experiment, all participants familiarized themselves with the gestural control and the KT for approximately $1.5 - 2$ minutes. Whereas in the end, each participant has filled up the User Experience Questionnaire (UEQ) \cite{Hinderks2019} and a custom made questionnaire asking familiarity with the devices, and whether they consider useful to keep on developing this project. 

The experiment consists of four tasks, and it is imposed a \textit{soft} limit of $5$ minutes to complete them. The temporal constraint limits the experiment duration to prevent fatigue and distractions that may affect the final result. The experiment is composed of four ordered tasks. When a volunteer completes a task, an experimenter evaluates if the remaining time is sufficient for the next one. If the time is not enough, the experimenter stops the volunteer, and the experiment ends. The experimenter does not interrupt the volunteer task execution; therefore, some experiments can exceed the soft limit. The description of the ordered tasks follows:

\begin{enumerate}
\item Starting from the main menu, the participant must select the \textit{playback} option, and reproduce the pre-recorded action \mbox{$\textsc{Move: A} \rightarrow \textsc{B}$}, which commands the robot to move the cube from position $A$ to position $B$.

\item Each participant must select the \textit{record} option from the main menu to record a new task, i.e., \mbox{$\textsc{Move: B} \rightarrow \textsc{A}$}, to bring the cube back from $B$ to $A$. Using KT, enabled by cuffing Baxter's wrist, the volunteer physically guides the robot through the task. Once the recording is finished, the wooden block is back to position $A$.

\item The third task consists in selecting the \textit{macro} option from the main menu and associating G1 and G2 (see Figure \ref{fig:gestures}), or two buttons while using the touchscreen, respectively to the pre-recorded \mbox{$\textsc{Move: A} \rightarrow \textsc{B}$} and the new \mbox{$\textsc{Move: B} \rightarrow \textsc{A}$}. After selecting \textit{play}, see Figure \ref{fig:HRI_macro}, the participant has to activate the two actions, using the corresponding inputs, to move the wooden block from $A$ to $B$ and \textit{vice versa}.

\item The last task consists in selecting the \textit{sequence} option from the main menu to create a sequence of actions and then selecting \textit{play} to start the reproduction. The sequence should include the prerecorded tasks \mbox{\textsc{action 1}} and \mbox{\textsc{action 2}}.
\end{enumerate}

Notice that to preserve a fair comparison, the experiment design does not include factors that could disadvantage the touchscreen interface, e.g., the need to use working gloves or hold a tool.

\begin{figure}[t]
	\centering
	\setlength\figureheight{0.49\linewidth}
	\setlength\figurewidth{0.49\linewidth}
	\footnotesize \input{images/confusion_matrix_online.tex}
	\caption[Confusion matrix for SLOTH online testing.]{Confusion matrix for online tests. The bottom row reports the \textit{recall} measures, while the rightmost column reports the \textit{precision} measures. The blue cell reports the overall accuracy.}
	\label{fig:confusion_matrix_online}
\end{figure}

\section{Results}
\label{sec:result}

This Section presents the SLOTH performances in our architecture and the results of the comparative study. We divide the study results between the evaluation of questionnaire answers and the analysis of the time needed to complete the experiment.

\subsection{Assessment of SLOTH Performance}
\label{sec:sloth}

To assess the performance of SLOTH and decide if retraining the model, we asked eleven volunteers to perform five repetitions for all the four gestures considered in our experiment, see Figure \ref{fig:gestures}. We conducted the assessment using our Android infrastructure streaming IMU data at $10$ $Hz$, as in the original work, directly to a workstation running SLOTH. The results are presented in the confusion matrix, Figure \ref{fig:confusion_matrix_online}. In general, the overall classification performance seems adequate for the study. In particular, the high level in precision indicates a reliable system that does not misinterpret gestures, preserving the correct association between gestures and commands, e.g., when the user performs G1, the system does not confuse it with G2, G3 or G4. However, the low recall levels suggest that user gestures may go undetected, e.g. when the user performs G1 has a 60\% chance that is not detected. 
We expect this to negatively affect the user experience since the user will have to perform multiple gesture repetitions for a single command.

\subsection{Analysis of Questionnaires}
The UEQ \cite{Hinderks2019}, the questionnaire used during the experiments, uses 26 questions to assess different properties related to user experience. The questions are grouped into six user experience scales: i) \textit{attractiveness}, i.e., the overall impression about the interface; ii) \textit{perspicuity}, i.e., the easiness of learning how to use the interface; iii) \textit{efficiency}, i.e., the utility of the interface in completing the task; iv) \textit{dependability}, i.e., the perceived security and control over the interface; v) \textit{stimulation}, i.e., the engagement generated by the interface; and vi) \textit{novelty}, i.e., the perceived novelty and aroused interest. Five of these scales can then be grouped in \textit{pragmatic qualities}, describing the task-related aspects of perspicuity, efficiency and dependability, and \textit{hedonic qualities}, including non-task-related aspects, such as stimulation and novelty. Since all the volunteers were from Italy, we used the Italian version of the questionnaire. We analysed the questionnaire answers using the Microsoft Excel sheet provided together with the questionnaire. The tool gets the volunteers answers and calculates scaled values. 


As a first step, we checked the value of the \emph{Cronbach\textquotesingle s alpha coefficient}, which provides a measure of internal consistency, i.e., how closely related a set of items are as a group. The rule-of-thumb is that the alpha coefficient should be higher or equal than $0.7$ to ensure scale reliability. Questionnaires related to the touchscreen interface have low alpha values only for dependency (0.35). Instead, for the gesture-based interface, we found low Cronbach's alpha for efficiency (0.55) and dependability (0.62). These findings can suggest a common misunderstanding of dependability items, but most probably are also influenced by the limited number of participants.

\label{sec:questionnaire}
\begin{figure*}[t]
    \centering
    \vspace{0.2cm}
    \includegraphics[trim={0.2cm 0.1cm 0.2cm 0.2cm},clip,width =0.96\linewidth]{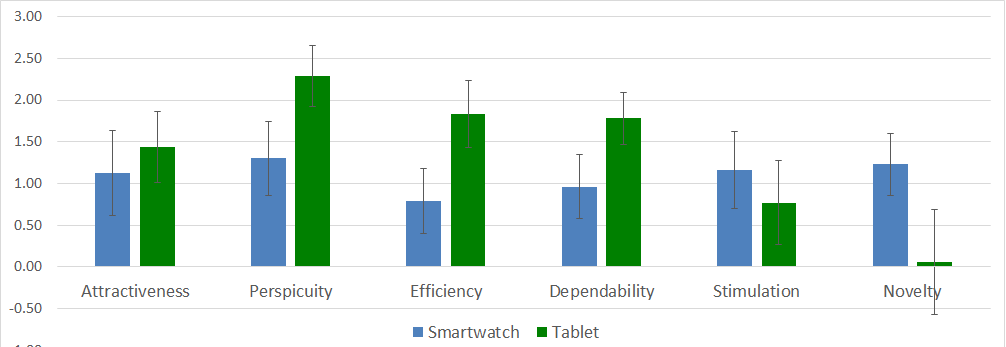}
    \caption{Comparison between gesture-based (in blue) and touchscreen-based (in green) experiments, related to usability and user experience aspects.}
    \label{fig:05_questionnaires}
\end{figure*}


Figure \ref{fig:05_questionnaires} presents the comparative results between the touchscreen interface (tablet, in green) and the gesture-based one (smartwatch, in blue). The bar graph represents the mean for each considered user experience scale. According to the UEQ guidelines, mean values between -0.8 and 0.8 represent a neutral evaluation of the corresponding scale, values greater than 0.8 represent a positive evaluation, and values less than -0.8 represent a negative evaluation. Instead, error bars represent the $95\%$ confidence of scale means, i.e., the interval whereby $95\%$ of the scale means are located if the experiment is repeated an infinite number of times. While comparing the values, if the two confidence intervals do not overlap, the difference is significant. Observing Figure \ref{fig:05_questionnaires}, a relevant difference is evident for pragmatic qualities (i.e., perspicuity, efficiency and dependability) where the tablet achieves higher evaluations. Instead, for hedonic qualities (i.e., stimulation and novelty), volunteers evaluated more positively the gesture-based interface. 

Finally, from the custom made questionnaire, we found out that $64\%$ of the participants were familiar with tablets, while only $12\%$ with smartwatches. Moreover, while tablet usage assumes a touchscreen interaction, smartwatches are not typically used for gestural interaction. This difference, together with the non-optimal performances of the gesture recognition system, probably justifies the disadvantage of gestural interaction in the pragmatic scales and its advantage in the hedonic ones.
 
\begin{figure}[t]
    \centering
    \includegraphics[trim={1cm 0 2cm  0},clip,width=0.65\linewidth]{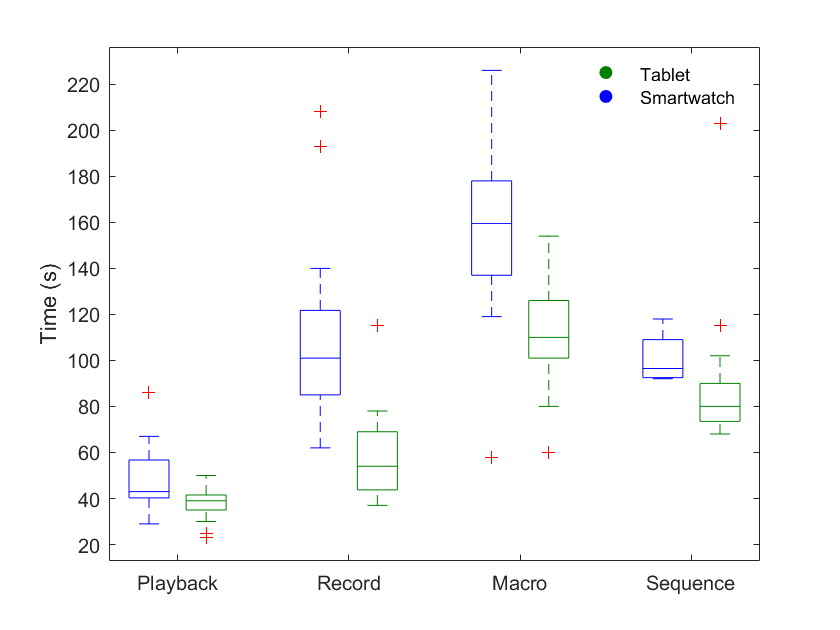}
    \caption{Box plots of the time that each participant needed to complete the four experimental tasks. In green data about the touchscreen-based interaction, in blue data for the gesture-based one. We reported outliers in red.}
    \label{fig:time}
\end{figure}

\subsection{Time Analysis}
\label{sec:timing}
Out of the 25 participants, only 5 completed the four tasks in the given time using the smartwatch, while 23 completed all the tasks using the tablet. This result is coherent with the participants' perception measured by the questionnaire about the overall interface efficiency. Figure \ref{fig:time} reports with box plots the statistics of the time spent to perform each of the four tasks with the tablet (in green) and the smartwatch (in blue). For both interfaces, we highlighted outliers in red. We point out that the box plots represent only the data of 23 volunteers since the subdivision, over the four tasks for the first two volunteers, have not been recorded. Furthermore, data for the sequence task using the gesture-based interface were available only for five participants since the others did not complete this task. 

The measured times do not measure only the user interaction but also include robot motions. This factor partially explains the higher variance for both touchscreen and smartwatch in the record and macro tasks. In fact, in these tasks, the time taken by the robot motion is different for each participant, since the \mbox{$\textsc{Move: B} \rightarrow \textsc{A}$} task is recorded individually by each of them. Instead, in the playback and the sequence tasks, the used robot motions, pre-recorded by the experimenters, are the same for everyone, and therefore the variance is comparatively lower. 

The playback task necessitates a few user inputs (i.e., selecting the playback option and then the motion to execute), and therefore the temporal difference between the two input modalities is limited. Instead, when the task becomes more complex (i.e., more user inputs are required to reach the end), the difference between the two input modalities increases due to the intrinsic difference between signals and events. This observation finds evidence in the timing results for the record and macro task but not for the equally complex sequence task. This result can be explained by observing that only five top-performing subjects reached this task using the smartwatch, and thus it suggests that the influence of volunteer skills is relevant. Therefore, more extensive user training could smooth the differences between gesture- and touchscreen-based interfaces.

\section{Conclusions}
\label{sec:conclusion}

In this work, we introduced a few designs principles for architectures supporting multiple human-robot interaction modalities. We presented the implementation of these design principles into an architecture mediating the human-robot interaction using both a smartwatch and a tablet. We evaluated the differences between the two communication modalities with a comparative experiment in a human-robot collaboration scenario.

The result of the experiments shows a positive evaluation for both interaction modalities  [$H_1$ and $H_2$]. The smartwatch and tablet interface received better evaluations, respectively, on the hedonistic [$H_{1.1}$] and pragmatic scales [$H_{2.1}$]. These results can be justified by the non-optimal performances of the gesture recognition system and the volunteers' familiarity with touchscreen technologies. We also observed the user proficiency effect in the task completion time, where the most skilled volunteers obtained similar results with the two input modalities.

These are not conclusive results because of the experimental limitations: the small number of subjects, simple collaborative scenario and lack of extensive statistical analysis. However, they encourage more in-depth analyses and studies of alternative interfaces in the HRC context. Future studies in this field should aim to improve the technical performances of gesture-based technologies, consider more complex experimental setups and investigate new interaction modalities.

\section*{Acknowledgements}
The authors would like to thank the teachers and students of the vocational education and training schools Centro Oratorio Votivo, Casa di Carit\`{a}, Arti e Mestieri di Ovada for their contribution to the experiments.
This work has been partially supported by the European Union Erasmus+ Programme via the European Master on Advanced Robotics Plus (EMARO+) program (grant agreement n. 2014-2616).

\bibliographystyle{splncs}
\bibliography{bibliography}

\end{document}